%% file: iclr2023_conference.tex
\documentclass{article} 
\usepackage{iclr2023_conference,times}
\input{math_commands.tex}

\usepackage{hyperref}
\usepackage{url}
\usepackage{graphicx}
\usepackage{float} 


\title{Empirical Analysis of the Strengths and Weaknesses of PEFT Techniques for LLMs}

\author{George Pu, Anirudh Jain, Jihan Yin \& Russell Kaplan \\
Scale AI\\
\texttt{\{george.pu,anirudh.jain,jihan.yin,russell.kaplan\}@scale.com}
}

\iclrfinalcopy 
\begin{document}

\maketitle

\begin{abstract}
As foundation models continue to exponentially scale in size, efficient methods of adaptation become increasingly critical. Parameter-efficient fine-tuning (PEFT), a recent class of techniques that require only modifying a small percentage of the model parameters, is currently the most popular method for adapting large language models (LLMs). Several PEFT techniques have recently been proposed with varying tradeoffs. We provide a comprehensive and uniform benchmark of various PEFT techniques across a representative LLM, the FLAN-T5 model, and evaluate model performance across different data scales of classification and generation datasets. Based on this, we provide a framework for choosing the optimal fine-tuning techniques given the task type and data availability. Contrary to popular belief, we also empirically prove that PEFT techniques converge slower than full tuning in low data scenarios, and posit the amount of data required for PEFT methods to both perform well and converge efficiently. Lastly, we further optimize these PEFT techniques by selectively choosing which parts of the model to train, and find that these techniques can be applied with significantly fewer parameters while maintaining and even improving performance.
\end{abstract}

\section{Introduction}
As large language models become widely adopted, efficient training and deployment become critical requirements for enabling widespread usage. Each task that an LLM is fine-tuned on requires an entirely different set of weights. When models scale to hundreds of billions of parameters, hosting a different set of weights for each model becomes widely inefficient and cost prohibitive while reloading all the weights for different tasks is too slow. Parameter-efficient fine-tuning techniques aim to solve this problem by modifying a very small portion of weights relative to the full model size while keeping the rest of the model frozen \citep{mao2021unipelt}.

At inference time, many adaptations of the same model can be served together by quickly swapping tiny submodules rather than all the weights. The current landscape of PEFT techniques is rapidly evolving and several PEFT techniques have recently been proposed -- each claiming to have advantages over the others in varying capacities. However, given that these techniques have each been evaluated in a silo on different models and datasets, it is unclear when to appropriately utilize one technique over another. This work seeks to provide a framework for evaluating how to effectively utilize PEFT by empirically evaluating which technique works well in what task types and how these techniques scale with data. Further, through an ablation study, our work seeks to understand which parts of the model are most important to train for a given task type and technique, leading to even more efficient adaptation and reduced parameter count. Our key contributions are: 

\begin{enumerate}
    \item Conducting a thorough comparison and analysis of the current state-of-the-art PEFT methods on the FLAN-T5 model across different data sizes and task types (generation/classification), evaluating a variety of dimensions including accuracy, convergence speeds and other relevant metrics. 
    \item Performing ablation studies to better understand the relative importance of updating various parts of the model when adapting LLMs, considering layer ordering and submodule granularities, and further optimizing PEFT techniques to reduce the number of trained parameters and ultimately improve efficiency.  
\end{enumerate}

\begin{figure}[t!]
\begin{center}
\includegraphics[width=6cm]{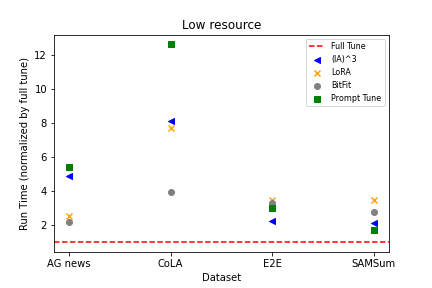}
\includegraphics[width=6cm]{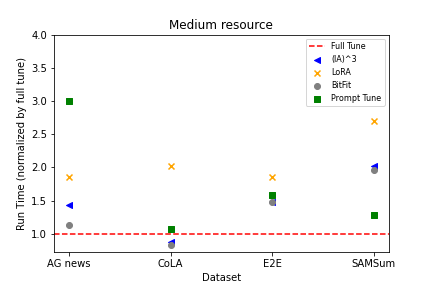}
\end{center}
\caption{Total run time of experiments normalized (PEFT time / full tune time) with different PEFT techniques across both low-resource (left) and medium-resource (right) scenarios. We limit high-resource experiments to one epoch, thus exclude these results. Also, we perform early stopping in all experiments where the stop criteria is defined by validation loss no longer decreasing.}
\label{peft-runtime}
\end{figure}

\section{Background and Methodology}
We provide background on large language models and parameter-efficient fine-tuning in appendix \ref{background} and detail the principles guiding our selection of methods and models in appendix \ref{design_choices}. Our experimentation centers on the FLAN-T5-XL model, and we explore the efficacy of four different fine-tuning techniques - LoRA, $(IA)^3$, prompt tuning, and BitFit - and compare their performance against a fully fine-tuned model trained on identical train/val/test splits. We conduct a comprehensive evaluation of the parameter-efficient fine-tuning (PEFT) methods and establish a framework to facilitate selection of the most appropriate technique in any given scenario.

To further optimize and improve upon existing PEFT techniques, we investigate which parts of the model are most important during fine-tuning. Compared to BitFit and prompt tuning, we conduct an ablation study with LoRA and $(IA)^3$ due to their configuration flexibility (e.g. attention blocks, dense layers, modified transformer layers). We analyze the effects of LoRA and $(IA)^3$ across which layers to apply the PEFT technique (e.g. early vs later vs random) and the impact of dropping out specific submodules, such as attention vectors and layer activations.

\section{Experiments}
\subsection{Experiment Setup}
Emulating current industry scenarios, we decided to evaluate the models against a variety of classification and generation datasets and a variety of data scales. For classification, we select AG news \citep{zhang2015character}, which has four classes and over 100,000 samples, and CoLA \citep{warstadt2019neural}, which has two classes and around 10,000 samples. For generation, we select the E2E dataset \citep{novikova2017e2e}, which is in the restaurant domain with 50,000 samples, and SAMSum \citep{gliwa2019samsum} for abstractive dialogue summarization with around 15,000 samples. To unify these experiments across datasets, we select three data scales: \textbf{low-resource} (at most 100 data points), \textbf{medium-resource} (at most 1,000 data points), and \textbf{high-resource} (at most 10,000 data points). Hyperparameters and implementation specifics are listed in appendix \ref{hyperparameters_implementation_specifics}.

\subsection{Analysis of PEFT Techniques}
\label{analysis-peft}

\begin{table}[t]
\caption{Benchmarking FLAN-T5 across data splits, measuring accuracy (exact string-match) for AG News/CoLA and ROUGE-L (longest common subsequence) for E2E NLG/SAMSum, where higher values are better.}
\label{benchmark-table}
\begin{center}
\begin{tabular}{lllll}
\multicolumn{1}{c}{\bf PEFT/Dataset}  &\multicolumn{1}{c}{AG News} &\multicolumn{1}{c}{CoLA} &\multicolumn{1}{c}{E2E NLG} &\multicolumn{1}{c}{SAMSum}
\\ \hline \\
Full Tuning (low)  &0.8588 &0.699 &0.46464 &0.3876\\
$(IA)^3$ (low) &0.67 &0.6973 &0.29508 &0.32924\\
LoRA (low) &0.8612 &\textbf{0.76436} &0.48257 &0.41197\\
BitFit (low) &\textbf{0.8808} &0.7203 &\textbf{0.48825} &\textbf{0.41914}\\
Prompt Tune (low) &0.60684 &0.68199 &0.0258 &0.00472\\
\\ \hline \\
Full Tuning (med.)  &\textbf{0.9212} &0.79119 &0.46523 &0.40908\\
$(IA)^3$ (med.) &0.9208 &0.78161 &0.48483 &0.43183\\
LoRA (med.) &0.9148 &\textbf{0.81418} &0.48364 &\textbf{0.43283}\\
BitFit (med.) &0.9156 &0.78736 &\textbf{0.48539} &0.42822\\
Prompt Tune (med.) &0.7312 &0.76245 &0.44173 &0.3792\\
\\ \hline \\
Full Tuning (high) &\textbf{0.934} &0.81417 &\textbf{0.48051} &0.43356\\
$(IA)^3$ (high) &0.9252 &0.80842 &0.4789 &\textbf{0.43998}\\
LoRA (high) &0.9264 &\textbf{0.83333} &0.4756 &0.43485\\
BitFit (high) &0.9192 &0.8295 &0.47973 &0.43098\\
Prompt Tune (high) &0.872 &0.80567 &0.45942 &0.3914
\end{tabular}
\end{center}
\end{table}

\textbf{PEFT techniques are slower to converge than full tuning in low/medium-resource scenarios.} In Figure~\ref{peft-runtime}, we surprisingly observe that full tuning consistently converges faster than PEFT techniques and results in higher model performance, except for with BitFit on CoLA in the medium-resource split. In low-resource, we observe that full tuning has a speedup of 73\% on AG News, 87\% on CoLA, 66\% on E2E NLG, and 60\% on SAMSum. In medium-resource, we observe that full tuning has a speedup of 46\% on AG News, 16\% on CoLA, 37\% on E2E NLG, and 64\% on SAMSum. Specific values are listed in Table~\ref{convergence-table}. These results suggest that for lower-resource datasets, if we prioritize training speed and less on hardware constraints, full tuning is a better option. However, it is interesting that most PEFT methods actually converge faster when there is more data available. We hypothesize that this is because, at lower data quantities, full tuning quickly learns and overfits to the smaller dataset while PEFT methods learn unstably, whereas, at higher data sizes, PEFT methods are more stable and better learn the underlying data structure.

\textbf{We benchmark a mixed bag of results where no obvious PEFT method performs best, but given data volume, there is a clear decision framework between full tuning and PEFT.} From Table~\ref{benchmark-table}, no optimal fine-tuning method exists based on the task, but there are specific scenarios where one method is much better. For example, BitFit and LoRA perform the best in low/medium-resource scenarios, and full tuning increases in relative performance as we increase the amount of data to higher samples. There exists a clear distinction between speed and performance, where PEFT has worse speed but better performance in low-resource and the inverse holds as data scales. 

Given the mixed results, we provide additional analysis on LoRA, $(IA)^3$ and BitFit across time and space dimensions to provide a holistic framework for choosing the optimal fine-tuning method. In addition, we benchmark these results against the full tuning baselines. Table~\ref{benchmark-parameters-table} gives a sense of what is the most memory-efficient method during training and for downstream applications. Table~\ref{benchmark-runtime-table} gives a sense of what is the most cost-efficient method with respect to hardware consumption. We calculate the values by dividing the performance results in Table~\ref{benchmark-table} by either the number of tunable parameters or total run time in seconds. 

To summarize our findings for different resource-constrained scenarios, we recommend full tuning across low/medium data splits and PEFT techniques in high-resource for time-constrained scenarios. For memory-constrained scenarios, we observe that BitFit and $(IA)^3$ perform best in low-resource scenarios, and $(IA)^3$ performs best in medium/high-resource scenarios. For performance-constrained scenarios, we recommend $(IA)^3$, LoRA or BitFit for low/medium-resource and full tuning in high-resource. In addition, $(IA)^3$ uses element-wise matrix multiplication and has the least memory overhead of the other methods because we can multiply the model weights with $(IA)^3$ vectors so there is no need to add or store additional parameters. Lower-resource scenarios are most common in production use cases, and if the amount of data samples is very small, prompt engineering or in-context learning are viable alternatives.

\subsection{Ablation Results}
\label{ablation-peft}
\begin{table}[t]
\caption{Ablations across specific subcomponents for discovering important factors towards performance on downstream tasks. Accuracy for classification and ROUGE-L for generation, where higher values are better.}
\label{ablation-table}
\begin{center}
\begin{tabular}{lllll}
\multicolumn{1}{c}{\bf PEFT/Dataset}  &\multicolumn{1}{c}{AG News} &\multicolumn{1}{c}{CoLA} &\multicolumn{1}{c}{E2E NLG} &\multicolumn{1}{c}{SAMSum}
\\ \hline \\
$(IA)^3$ (all layers) &0.9208 &0.78161 &0.48483 &0.43183 \\
LoRA (all layers) &0.9148 &0.814176 &0.48364 &0.43283 \\
$(IA)^3$ (select early layers) &0.60 &0.6877 &0.462 &0.4154\\
LoRA (select early layers) &0.9104 &0.70306 &0.4823 &\textbf{0.43462} \\
$(IA)^3$ (select later layers) &\textbf{0.9212} &0.7796 &0.48550 &0.4128 \\
LoRA (select later layers) &0.8892 &\textbf{0.82183} &\textbf{0.48553} &0.43028 \\
$(IA)^3$ (select 50\% random layers) &0.918 &0.74712 &0.40 &0.416 \\
LoRA (select 50\% random layers) &0.9096 &0.78735 &0.4834 &0.43109 \\
LoRA (drop self-attention) &0.8928 &0.7107 &0.48055 &0.415 \\
LoRA (drop enc/dec-attention) &0.9164 &0.8007 &0.47966 &0.43218 \\
LoRA (drop query/output attention) &0.9152 &0.8084 &0.4815 &0.43459 \\
$(IA)^3$ (drop attention) &0.6972 &0.7413 &0.40 &0.2359 \\
$(IA)^3$ (drop dense activation) &0.918 &0.7643 &0.481544 &0.4292 \\
\end{tabular}
\end{center}
\end{table}
\textbf{Performance degradation with selective module adaptation affects $(IA)^3$ more, whereas LoRA is more robust.} We are able to show that parameter count can be reduced by 50\% while maintaining performance, leading to a more efficient and adaptable model. Interestingly enough, our ablation results have higher downstream performance than the LoRA or $(IA)^3$ applied to all layers. Performance degradation is quite negligible with reduced parameters, showing that we can further reduce the number of parameters to less than the original techniques suggested (e.g. PEFT random or later layers).

\textbf{Attention is critical in classification and generation tasks and is especially true for $(IA)^3$ compared to LoRA.} Dropping self-attention hurts LoRA in classification tasks more compared to removing query/output vectors and encoder-decoder attention. For models that performed equivalent to the full technique, we were able to reduce parameters by at least half and include the parameter counts per variation in Table~\ref{ablation-param-table}.

\textbf{Adapting earlier layers does poorly compared to later or random layers in $(IA)^3$.} In the later or random layer selection, performance is much better in $(IA)^3$ compared to modifying early layers. The ablation that performs best with the minimum number of parameters is selecting and PEFT-ing the later layers. These findings align with the importance of modifying later representations, similar to transfer learning paradigms.

\section{Conclusion}
In this paper, we benchmark LLMs across several dimensions, develop an understanding of where LLM PEFT techniques should be adapted, and provide a framework on choosing the optimal technique. On our baselines, we find that PEFT methods generally perform better on low/medium-resource levels compared to fully supervised fine-tuning, but are slower to converge. On our ablations, we discover the importance of attention-level modifications and selecting later layers in downstream performance. We can further prune and optimize LoRA to even smaller degrees with module/layer-specific granularities, increasing its adaptability without sacrificing performance.

No method is strictly better, but we aim to establish empirical guidelines for which fine-tuning methods to use for downstream tasks based on recommendations in \ref{analysis-peft} and \ref{ablation-peft}. The efficiency-performance tradeoff depends on each use case, but we empirically prove that there are significant parameter reductions that can be made on top of the original PEFT techniques without impacting model performance. This study provides a comprehensive analysis of PEFT methods and further optimization opportunities to make applying these techniques more systematic and approachable. 

\bibliography{iclr2023_conference}
\bibliographystyle{iclr2023_conference}

\newpage
\appendix
\section{Appendix}

\subsection{Background and Related Work}
\label{background}
\textbf{Large language models (LLMs)} have exploded in popularity recently, across many different Transformer architectures and optimization techniques. Within the LLM landscape, models are either discriminative or generative, where the former learns the decision boundaries and the latter models the distribution of classes. \citet{devlin2018bert} introduced BERT, a discriminative model, comprised of encoder networks optimized on masked language modeling and next sentence prediction. Other variants are RoBERTA and ALBERT \citep{liu2019roberta, lan2019albert}. Generative models also exist for decoder-only variants like GPT \citep{brown2020language}, GPT-neo \citep{black2022gpt}, and BLOOM \citep{scao2022bloom}, and encoder-decoder variants such as T5 and T0 \citep{raffel2020exploring, sanh2021multitask}. Instruction fine-tuning has also garnered recent interest, being applied in FLAN-T5 and FLAN-PaLM \citep{wei2021finetuned, chung2022scaling}. However, these models are comprised of billions of parameters, making them costly to train and serve.

\textbf{Parameter-Efficient Fine-Tuning (PEFT)} For more efficient training and serving, PEFT techniques are growing in popularity due to their ability to be rapidly adapted and trained. Within the PEFT landscape, there are several architectural design choices on where the additional (sub)modules should be attached to the pre-trained language model, which varies on modifying either the underlying model, features and parameters. Some recent work includes adapter layers, which modify the transformer layer by adding new feedforward layers that comprise of a down-projection, non-linearity and up-projection \citep{houlsby2019parameter}. On the feature level, \citet{lester2021power} introduce prompt tuning, which prepends soft prompts to the input text. On the parameter level, \citet{hu2021lora} introduce LoRA, which adapts the attention weights (query, key, value, and output vector). Similarly, $(IA)^3$ modifies the key and value attention weights along with feedforward activation with element-wise multiplication \citep{liu2022few}. 

These methods generally involve freezing the underlying model, but other techniques arise in which only a portion of the LLM is changed. For example, \citet{zaken2021bitfit} hypothesize that fine-tuning exposes knowledge induced by LM training and introduce BitFit, which fine-tunes the bias terms of the model. Prior work also demonstrates the importance and robustness of adapter-based tuning during fewer data scenarios, while being less sensitive to learning rate \citep{he2021effectiveness}. In addition, \citet{chen2022revisiting} performed a PEFT evaluation with RoBERTa and found better performance of these techniques on fewer data tasks, the importance of fine-tuning and pitfalls of prefix tuning. We aim to extend these findings with larger models (from hundreds of millions to billions of parameters) and incorporate $(IA)^3$ within our benchmark.

\subsection{Design Choices}
\label{design_choices}

Design choices were based on industry-needs and common practices for deploying parameter-efficient fine-tuning strategies with DeepSpeed inference. We wanted to explore PEFT methods while exploring breadth of architectural changes and maintaining depth in evaluating to understand the respective method/effects across our downstream use cases. During our model exploration period, we ran initial experiments across various model sizes along with autoregressive models, such as GPT-Neo, BLOOM and T5 variants. We found noticeably worse performance with other models compared to FLAN-T5 across all datasets and metrics, so we choose FLAN-T5 as a representative LLM. 

Due to compute and cost constraints, we were unable to perform our experiments across the entire set of model architectures and sample variance across multiple trials. However, we were able to validate the small impact of variance attributable across our initial experiments and encoder-decoder and decoder architectures, such as GPT-Neo and T5 variants. Empirically, when we ran these experiments twice on different random seeds, the convergence and performance trends were similar. Also, we ensure the same val/test sets with a random seed value and cap the size to 2500 samples across both splits. To maintain similar step updates, we limit low-resource to 10 epochs, medium-resource to 5 epochs and high-resource to one epoch while also incorporating early-stopping with model loss on the validation set. Given the randomized data splits, some dataset benchmarks may include mixed results due to randomness, but we ensure all methods are evaluated on the same splits. 

We implement LoRA, $(IA)^3$, prompt tuning and BitFit and benchmark against a fully fine-tuned model on the same train/val/test splits. All these PEFT techniques modify less than 1\% of the model's parameters, and due to memory constraints, we use a fixed set of configurations for each PEFT method and model hyperparameters based on best practices. We choose these PEFT techniques because they focus on different levels of architecture adaptation, such as at the parameter level with LoRA and $(IA)^3$, feature level with prompt tuning, and model level with BitFit. Implementation specifics are in the next section. On the practical side, $(IA)^3$ is easier to implement than BitFit for most models due to optimization libraries like DeepSpeed fusing inter-transformation operations together \citep{rasley2020deepspeed}, making it harder to decouple the bias term. 

\subsection{Hyperparameters and Implementation Specifics}
\label{hyperparameters_implementation_specifics}

We use the HuggingFace Transformer's package to implement our FLAN-T5-XL model, along with initially evaluating model sizes and decoder architectures with the GPT-Neo and GPT-J models. For optimization, we use the LambdaLR learning rate scheduler with AdamW optimizer, which requires specifying an initial, warmup and linear annealing learning rate. In addition, we use 16 batch size with a gradient accumulation size of 4. We run all of our experiments with 4 A10s using model parallelism.

For the PEFT hyperparameters, we follow prior art in configuring the default hyperparameters across model architectures and PEFT techniques. Due to time and memory constraints, we use one fixed set of hyperparameters for all our low/medium/high-resource experiments across dataset types. For prompt tuning, the authors found 20-100 prompt tokens worked best, and we opted for 100 during adaptation. For LoRA, we follow the original paper with a dimension rank of 2, and $(IA)^3$ initializes the scaling term to one. BitFit is dependent on the specific layer's output features and is initialized as a tensor of zeros. In prompt tuning, we modify the input embeddings to support continuous/soft embeddings prepended to the input text. In LoRA and $(IA)^3$, we follow \citet{liu2022few} in morphing the LoRA matrix-multiplication adaptation in attention blocks to also support element-wise scaling for $(IA)^3$ to rescale the key and values in attention mechanisms. We differentiate the submodules, such as T5's "EncDecAttention", "SelfAttention" and "DenseReluDense" blocks. For BitFit, we adapt all bias terms in the linear layers to require gradients.

Given our compute and cost constraints, we use hyperparameters from prior work, which have all performed hyperparameter search for optimal and general values. More information can be found in the FLAN-T5, prompt-tuning, $(IA)^3$, LoRA and BitFit papers in the references. For full fine-tuning, we set the initial LR to $3e-5$, warmup LR to $3e-4$, and annealing LR to $3e-5$. For LoRA, $(IA)^3$ and BitFit, we set the initial LR to $3e-4$, warmup LR to $3e-3$, and annealing LR to $3e-4$. For prompt tuning, we set the initial LR to $3e-3$, warmup LR to $3e-1$, and annealing LR to $3e-2$.

\subsection{Parameter Counts Of PEFT Techniques and Ablation Study}
For the parameter count of baselines, we use the FLAN-T5-XL model, which has 2,849,757,184 model parameters. Given our implementation specifics, the $(IA)^3$ method has 933,888 parameters. The LoRA method has 3,538,944 parameters. Prompt tuning has 204,800 parameters and scales in n * h, where h is the hidden state size and n is the length of the prefix. Lastly, BitFit modifies 1,179,648 parameters. As a result, all these methods use a small fraction of the total parameter weights. Also, we include information on specific parameter counts on our ablation experiments in Table~\ref{ablation-param-table}, which vary based on the particular subcomponent modified (e.g. layers).

\begin{table}[t]
\caption{Parameter count across ablation variants, including $(IA)^3$ and LoRA applied on the full layers/modules.}
\label{ablation-param-table}
\begin{center}
\begin{tabular}{lllll}
\multicolumn{1}{c}{\bf Ablation Technique}  &\multicolumn{1}{c}{Parameter Count}
\\ \hline \\
Total Parameters (no PEFT) &2849757184\\
$(IA)^3$ (full layers/modules) &933888 \\
LoRA (full layers/modules) &3538944 \\
$(IA)^3$ (select early layers) &466944 \\
LoRA (select early layers) &1769472 \\
$(IA)^3$ (select later layers) &466944 \\
LoRA (select later layers) &1769472 \\
$(IA)^3$ (select random layers) &466944 \\
LoRA (select random layers) &1769472 \\
LoRA (drop self-attention) &1179648 \\
LoRA (drop enc/dec-attention) &2359296 \\
LoRA (drop query/output attention) &1769472 \\
$(IA)^3$ (drop attention) &344064 \\
$(IA)^3$ (drop dense activation) &589824 \\
\end{tabular}
\end{center}
\end{table}
\newpage
\subsection{Additional Results}
In our benchmark, we analyze different results that are mentioned in Section \ref{analysis-peft}. Table~\ref{convergence-table} refers to the exact numbers used in calculating exact convergence times from Figure~\ref{peft-runtime}. In addition, we supply the tables used in determining the fine-tuning framework dependent on the number of parameters used and total convergence time based on early stopping with validation loss. Given that prompt tuning performs noticeably worse than the other PEFT techniques, we exclude these results from comparisons. 

\begin{table}[H]
\caption{FLAN-T5 convergence table across epochs and total run time (in seconds), where lower values are better. Low refers to low-resource, whereas medium refers to medium-resource data splits. We limit high-resource experiments to one epoch, thus exclude these results.}
\label{convergence-table}
\begin{center}
\begin{tabular}{lllll}
\multicolumn{1}{c}{\bf PEFT/Dataset}  &\multicolumn{1}{c}{AG News} &\multicolumn{1}{c}{CoLA} &\multicolumn{1}{c}{E2E NLG} &\multicolumn{1}{c}{SAMSum}
\\ \hline \\
Full Tuning (low)  &\textbf{Ep. 4; 1249} &\textbf{Ep. 2; 89} &Ep. \textbf{3; 2845} &\textbf{Ep. 3; 1951}\\
$(IA)^3$ (low) &Ep. 9; 6130 &Ep. 9; 725 &Ep. 9; 6307 &Ep. 9; 4096\\
LoRA (low) &Ep. 5; 3146 &Ep. 6; 687 &Ep. 9; 9897 &Ep. 9; 6758\\
BitFit (low) &Ep. 8; 2738 &Ep. 7; 349 &Ep. 9; 9426 &Ep. 9; 5385\\
Prompt Tune (low) &Ep. 7; 6797 &Ep. 9; 1124 &Ep. 4; 8609 &Ep. 3; 3340\\
\\ \hline \\
Full Tuning (med.)  &\textbf{Ep. 2; 1782} &Ep. 2; 608 &\textbf{Ep. 3; 3877} &\textbf{Ep. 1; 2089}\\
$(IA)^3$ (med.) &Ep. 4; 2565 &Ep. 3; 537 &Ep. 4; 5731 &Ep. 3; 4234\\
LoRA (med.) &Ep. 3; 3303 &Ep. 3; 1231 &Ep. 3; 7185 &Ep. 4; 5627\\
BitFit (med.) &Ep. 3; 2027 &\textbf{Ep. 3; 507} &Ep. 4; 5712 &Ep. 4; 4101\\
Prompt Tune (med.) &Ep. 2; 5338 &Ep. 2; 652 &Ep. 3; 6152 &Ep. 2; 2681\\
\end{tabular}
\end{center}
\end{table}

\begin{table}[H]
\caption{Measuring the relation of accuracy and parameters tuned with each PEFT technique against our full tuning baselines. Performance (accuracy or ROUGE-L) is divided by the number of tunable parameters of each experiment, normalized to each hundred thousand parameters used.}
\label{benchmark-parameters-table}
\begin{center}
\begin{tabular}{lllll}
\multicolumn{1}{c}{\bf PEFT/Dataset}  &\multicolumn{1}{c}{AG News} &\multicolumn{1}{c}{CoLA} &\multicolumn{1}{c}{E2E NLG} &\multicolumn{1}{c}{SAMSum}
\\ \hline \\
Full Tuning (low) &3.01E-05 &2.45E-05 &1.63E-05 &1.36E-05\\
$(IA)^3$ (low) &0.0717 &\textbf{0.0746} &0.03159 &0.03525\\
LoRA (low) &0.0243 &0.0216 &0.0136 &0.01164\\
BitFit (low) &\textbf{0.0746} &0.06106 &\textbf{0.04139} &\textbf{0.03553}\\
\\ \hline \\
Full Tuning (med.) &3.23E-05 &2.77E-05 &1.63E-05 &1.44E-05\\
$(IA)^3$ (med.) &\textbf{0.0985} &\textbf{0.0836} &\textbf{0.05191} &\textbf{0.04624}\\
LoRA (med.) &0.0258 &0.023 &0.01366 &0.01223\\
BitFit (med.) &0.07761 &0.06674 &0.04118 &0.0363\\
\\ \hline \\
Full Tuning (high) &3.28E-05 &2.86E-05 &1.69E-05 &1.52E-05\\
$(IA)^3$ (high) &\textbf{0.0991} &\textbf{0.0865} &\textbf{0.05128} &\textbf{0.04711}\\
LoRA (high) &0.0262 &0.0235 &0.01343 &0.01228\\
BitFit (high) &0.0779 &0.0703 &0.04066 &0.03643\\

\end{tabular}
\end{center}
\end{table}

\begin{table}[H]
\caption{Measuring the relation of accuracy and total run time with each PEFT technique against our full tuning baselines. Performance (accuracy or ROUGE-L) is divided by the the total run time of each experiment (in minutes).}
\label{benchmark-runtime-table}
\begin{center}
\begin{tabular}{lllll}
\multicolumn{1}{c}{\bf PEFT/Dataset}  &\multicolumn{1}{c}{AG News} &\multicolumn{1}{c}{CoLA} &\multicolumn{1}{c}{E2E NLG} &\multicolumn{1}{c}{SAMSum}
\\ \hline \\
Full Tuning (low) &\textbf{0.04125} &\textbf{0.47124} &\textbf{0.0098} &\textbf{0.0119}\\
$(IA)^3$ (low) &0.00597 &0.05771 &0.0028 &0.00482\\
LoRA (low) &0.01592 &0.0583 &0.0029 &0.00365\\
BitFit (low) &0.0193 &0.12383 &0.0031 &0.00467\\
\\ \hline \\
Full Tuning (med.) &\textbf{0.03205} &0.07807 &\textbf{0.00719} &\textbf{0.01174}\\
$(IA)^3$ (med.) &0.02139 &0.08733 &0.00507 &0.00612\\
LoRA (med.) &0.01662 &0.03968 &0.00403 &0.00461\\
BitFit (med.) &0.0271 &\textbf{0.09317} &0.00509 &0.00626\\

\end{tabular}
\end{center}
\end{table}

\end{document}

%% file: math_commands.tex
\usepackage{amsmath,amsfonts,bm}

\def\eqref#1{equation~\ref{#1}}

\def\1{\bm{1}}

\DeclareMathAlphabet{\mathsfit}{\encodingdefault}{\sfdefault}{m}{sl}
\SetMathAlphabet{\mathsfit}{bold}{\encodingdefault}{\sfdefault}{bx}{n}













%% file: iclr2023_conference.bbl
\begin{thebibliography}{23}
\providecommand{\natexlab}[1]{#1}
\providecommand{\url}[1]{\texttt{#1}}
\expandafter\ifx\csname urlstyle\endcsname\relax
  \providecommand{\doi}[1]{doi: #1}\else
  \providecommand{\doi}{doi: \begingroup \urlstyle{rm}\Url}\fi

\bibitem[Black et~al.(2022)Black, Biderman, Hallahan, Anthony, Gao, Golding,
  He, Leahy, McDonell, Phang, et~al.]{black2022gpt}
Sid Black, Stella Biderman, Eric Hallahan, Quentin Anthony, Leo Gao, Laurence
  Golding, Horace He, Connor Leahy, Kyle McDonell, Jason Phang, et~al.
\newblock Gpt-neox-20b: An open-source autoregressive language model.
\newblock \emph{arXiv preprint arXiv:2204.06745}, 2022.

\bibitem[Brown et~al.(2020)Brown, Mann, Ryder, Subbiah, Kaplan, Dhariwal,
  Neelakantan, Shyam, Sastry, Askell, et~al.]{brown2020language}
Tom Brown, Benjamin Mann, Nick Ryder, Melanie Subbiah, Jared~D Kaplan, Prafulla
  Dhariwal, Arvind Neelakantan, Pranav Shyam, Girish Sastry, Amanda Askell,
  et~al.
\newblock Language models are few-shot learners.
\newblock \emph{Advances in neural information processing systems},
  33:\penalty0 1877--1901, 2020.

\bibitem[Chen et~al.(2022)Chen, Liu, Meng, and Liang]{chen2022revisiting}
Guanzheng Chen, Fangyu Liu, Zaiqiao Meng, and Shangsong Liang.
\newblock Revisiting parameter-efficient tuning: Are we really there yet?
\newblock \emph{arXiv preprint arXiv:2202.07962}, 2022.

\bibitem[Chung et~al.(2022)Chung, Hou, Longpre, Zoph, Tay, Fedus, Li, Wang,
  Dehghani, Brahma, et~al.]{chung2022scaling}
Hyung~Won Chung, Le~Hou, Shayne Longpre, Barret Zoph, Yi~Tay, William Fedus,
  Eric Li, Xuezhi Wang, Mostafa Dehghani, Siddhartha Brahma, et~al.
\newblock Scaling instruction-finetuned language models.
\newblock \emph{arXiv preprint arXiv:2210.11416}, 2022.

\bibitem[Devlin et~al.(2018)Devlin, Chang, Lee, and Toutanova]{devlin2018bert}
Jacob Devlin, Ming-Wei Chang, Kenton Lee, and Kristina Toutanova.
\newblock Bert: Pre-training of deep bidirectional transformers for language
  understanding.
\newblock \emph{arXiv preprint arXiv:1810.04805}, 2018.

\bibitem[Gliwa et~al.(2019)Gliwa, Mochol, Biesek, and Wawer]{gliwa2019samsum}
Bogdan Gliwa, Iwona Mochol, Maciej Biesek, and Aleksander Wawer.
\newblock Samsum corpus: A human-annotated dialogue dataset for abstractive
  summarization.
\newblock \emph{arXiv preprint arXiv:1911.12237}, 2019.

\bibitem[He et~al.(2021)He, Liu, Ye, Tan, Ding, Cheng, Low, Bing, and
  Si]{he2021effectiveness}
Ruidan He, Linlin Liu, Hai Ye, Qingyu Tan, Bosheng Ding, Liying Cheng, Jia-Wei
  Low, Lidong Bing, and Luo Si.
\newblock On the effectiveness of adapter-based tuning for pretrained language
  model adaptation.
\newblock \emph{arXiv preprint arXiv:2106.03164}, 2021.

\bibitem[Houlsby et~al.(2019)Houlsby, Giurgiu, Jastrzebski, Morrone,
  De~Laroussilhe, Gesmundo, Attariyan, and Gelly]{houlsby2019parameter}
Neil Houlsby, Andrei Giurgiu, Stanislaw Jastrzebski, Bruna Morrone, Quentin
  De~Laroussilhe, Andrea Gesmundo, Mona Attariyan, and Sylvain Gelly.
\newblock Parameter-efficient transfer learning for nlp.
\newblock In \emph{International Conference on Machine Learning}, pp.\
  2790--2799. PMLR, 2019.

\bibitem[Hu et~al.(2021)Hu, Shen, Wallis, Allen-Zhu, Li, Wang, Wang, and
  Chen]{hu2021lora}
Edward~J Hu, Yelong Shen, Phillip Wallis, Zeyuan Allen-Zhu, Yuanzhi Li, Shean
  Wang, Lu~Wang, and Weizhu Chen.
\newblock Lora: Low-rank adaptation of large language models.
\newblock \emph{arXiv preprint arXiv:2106.09685}, 2021.

\bibitem[Lan et~al.(2019)Lan, Chen, Goodman, Gimpel, Sharma, and
  Soricut]{lan2019albert}
Zhenzhong Lan, Mingda Chen, Sebastian Goodman, Kevin Gimpel, Piyush Sharma, and
  Radu Soricut.
\newblock Albert: A lite bert for self-supervised learning of language
  representations.
\newblock \emph{arXiv preprint arXiv:1909.11942}, 2019.

\bibitem[Lester et~al.(2021)Lester, Al-Rfou, and Constant]{lester2021power}
Brian Lester, Rami Al-Rfou, and Noah Constant.
\newblock The power of scale for parameter-efficient prompt tuning.
\newblock \emph{arXiv preprint arXiv:2104.08691}, 2021.

\bibitem[Liu et~al.(2022)Liu, Tam, Muqeeth, Mohta, Huang, Bansal, and
  Raffel]{liu2022few}
Haokun Liu, Derek Tam, Mohammed Muqeeth, Jay Mohta, Tenghao Huang, Mohit
  Bansal, and Colin Raffel.
\newblock Few-shot parameter-efficient fine-tuning is better and cheaper than
  in-context learning.
\newblock \emph{arXiv preprint arXiv:2205.05638}, 2022.

\bibitem[Liu et~al.(2019)Liu, Ott, Goyal, Du, Joshi, Chen, Levy, Lewis,
  Zettlemoyer, and Stoyanov]{liu2019roberta}
Yinhan Liu, Myle Ott, Naman Goyal, Jingfei Du, Mandar Joshi, Danqi Chen, Omer
  Levy, Mike Lewis, Luke Zettlemoyer, and Veselin Stoyanov.
\newblock Roberta: A robustly optimized bert pretraining approach.
\newblock \emph{arXiv preprint arXiv:1907.11692}, 2019.

\bibitem[Mao et~al.(2021)Mao, Mathias, Hou, Almahairi, Ma, Han, Yih, and
  Khabsa]{mao2021unipelt}
Yuning Mao, Lambert Mathias, Rui Hou, Amjad Almahairi, Hao Ma, Jiawei Han,
  Wen-tau Yih, and Madian Khabsa.
\newblock Unipelt: A unified framework for parameter-efficient language model
  tuning.
\newblock \emph{arXiv preprint arXiv:2110.07577}, 2021.

\bibitem[Novikova et~al.(2017)Novikova, Du{\v{s}}ek, and
  Rieser]{novikova2017e2e}
Jekaterina Novikova, Ond{\v{r}}ej Du{\v{s}}ek, and Verena Rieser.
\newblock The e2e dataset: New challenges for end-to-end generation.
\newblock \emph{arXiv preprint arXiv:1706.09254}, 2017.

\bibitem[Raffel et~al.(2020)Raffel, Shazeer, Roberts, Lee, Narang, Matena,
  Zhou, Li, and Liu]{raffel2020exploring}
Colin Raffel, Noam Shazeer, Adam Roberts, Katherine Lee, Sharan Narang, Michael
  Matena, Yanqi Zhou, Wei Li, and Peter~J Liu.
\newblock Exploring the limits of transfer learning with a unified text-to-text
  transformer.
\newblock \emph{The Journal of Machine Learning Research}, 21\penalty0
  (1):\penalty0 5485--5551, 2020.

\bibitem[Rasley et~al.(2020)Rasley, Rajbhandari, Ruwase, and
  He]{rasley2020deepspeed}
Jeff Rasley, Samyam Rajbhandari, Olatunji Ruwase, and Yuxiong He.
\newblock Deepspeed: System optimizations enable training deep learning models
  with over 100 billion parameters.
\newblock In \emph{Proceedings of the 26th ACM SIGKDD International Conference
  on Knowledge Discovery \& Data Mining}, pp.\  3505--3506, 2020.

\bibitem[Sanh et~al.(2021)Sanh, Webson, Raffel, Bach, Sutawika, Alyafeai,
  Chaffin, Stiegler, Scao, Raja, et~al.]{sanh2021multitask}
Victor Sanh, Albert Webson, Colin Raffel, Stephen~H Bach, Lintang Sutawika,
  Zaid Alyafeai, Antoine Chaffin, Arnaud Stiegler, Teven~Le Scao, Arun Raja,
  et~al.
\newblock Multitask prompted training enables zero-shot task generalization.
\newblock \emph{arXiv preprint arXiv:2110.08207}, 2021.

\bibitem[Scao et~al.(2022)Scao, Fan, Akiki, Pavlick, Ili{\'c}, Hesslow,
  Castagn{\'e}, Luccioni, Yvon, Gall{\'e}, et~al.]{scao2022bloom}
Teven~Le Scao, Angela Fan, Christopher Akiki, Ellie Pavlick, Suzana Ili{\'c},
  Daniel Hesslow, Roman Castagn{\'e}, Alexandra~Sasha Luccioni, Fran{\c{c}}ois
  Yvon, Matthias Gall{\'e}, et~al.
\newblock Bloom: A 176b-parameter open-access multilingual language model.
\newblock \emph{arXiv preprint arXiv:2211.05100}, 2022.

\bibitem[Warstadt et~al.(2019)Warstadt, Singh, and Bowman]{warstadt2019neural}
Alex Warstadt, Amanpreet Singh, and Samuel~R Bowman.
\newblock Neural network acceptability judgments.
\newblock \emph{Transactions of the Association for Computational Linguistics},
  7:\penalty0 625--641, 2019.

\bibitem[Wei et~al.(2021)Wei, Bosma, Zhao, Guu, Yu, Lester, Du, Dai, and
  Le]{wei2021finetuned}
Jason Wei, Maarten Bosma, Vincent~Y Zhao, Kelvin Guu, Adams~Wei Yu, Brian
  Lester, Nan Du, Andrew~M Dai, and Quoc~V Le.
\newblock Finetuned language models are zero-shot learners.
\newblock \emph{arXiv preprint arXiv:2109.01652}, 2021.

\bibitem[Zaken et~al.(2021)Zaken, Ravfogel, and Goldberg]{zaken2021bitfit}
Elad~Ben Zaken, Shauli Ravfogel, and Yoav Goldberg.
\newblock Bitfit: Simple parameter-efficient fine-tuning for transformer-based
  masked language-models.
\newblock \emph{arXiv preprint arXiv:2106.10199}, 2021.

\bibitem[Zhang et~al.(2015)Zhang, Zhao, and LeCun]{zhang2015character}
Xiang Zhang, Junbo Zhao, and Yann LeCun.
\newblock Character-level convolutional networks for text classification.
\newblock \emph{Advances in neural information processing systems}, 28, 2015.

\end{thebibliography}
